\documentclass{article}

\usepackage[utf8]{inputenc}
\usepackage[T1]{fontenc}
\usepackage{amsmath,amssymb,amsfonts}
\usepackage{graphicx}
\usepackage{booktabs}
\usepackage{hyperref}
\usepackage{cleveref}
\usepackage{algorithm}
\usepackage{algorithmic}
\usepackage{pifont}

\usepackage{tabularx}
\usepackage{array}
\usepackage{xcolor}
\usepackage{natbib}
\usepackage{tikz}
\usepackage{pgfplots}
\pgfplotsset{compat=1.18}
\usetikzlibrary{patterns}
\usepgfplotslibrary{fillbetween, groupplots}
\pgfplotsset{
  adaaxis/.style={
    width=0.6\linewidth, height=4.5cm,
    grid=major, grid style={gray!12},
    tick label style={font=\footnotesize}, label style={font=\footnotesize},
    legend style={font=\footnotesize, draw=none, fill=none},
    every axis plot/.append style={line width=1pt},
  },
}
\definecolor{adagold}{HTML}{C99700}
\definecolor{grpogray}{HTML}{8A8A8A}
\definecolor{prefblue}{HTML}{5B6BD0}
\definecolor{deadred}{HTML}{C1494A}
\definecolor{dialteal}{HTML}{2A9D8F}
\usepackage[margin=1in]{geometry}

\newcommand{\kg}{\ensuremath{k/G}}
\newcommand{\method}{AdaPrefix-GRPO}

\title{Max Out GRPO Signal:\\Adaptive Trace Prefix Control for Hard Reasoning Problems}
\author{Vladislav Beliaev\\
Independent Researcher\\
\texttt{belyaev.vladislav.nw@gmail.com}\\
\href{https://thinkdense.ai}{\texttt{thinkdense.ai}}}
\date{}

\begin{document}

\maketitle

\begin{abstract}
Group Relative Policy Optimization (GRPO) stalls on a model's hardest problems: when no rollout in a group succeeds, the group-relative advantages vanish and the problem contributes no gradient, wasting the frontier examples we most want to learn from. Prepending a correct prefix of a reference solution raises the success rate, making prefix length a continuous knob on difficulty. Concurrent methods set the knob once; \method{} turns it into a feedback controller: throughout training it adjusts how much of the solution each problem gets, holding its success rate near 50\%, where GRPO's gradient signal is largest, then withdraws the assistance entirely, so the deployed model solves problems unaided. On hard math, at matched training FLOPs, it more than doubles GRPO's accuracy on held-out problems from the training distribution for a 0.6B model (2.1$\times$), with 1.6$\times$ on Qwen3-1.7B and 1.7$\times$ on AIME, while roughly halving trace length. The method is implemented in data preparation plus a loss mask on prefix tokens; the trainer is otherwise stock. The smaller the model, the larger the gain.
\end{abstract}

\section{Introduction}
\label{sec:intro}

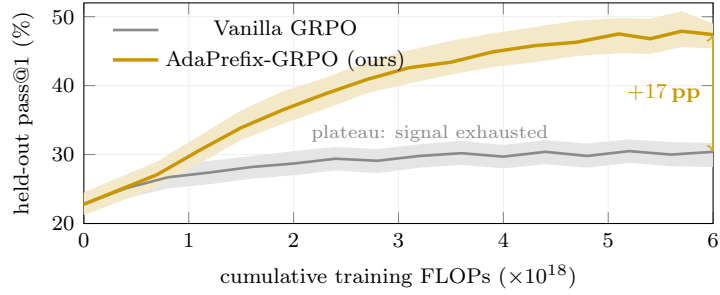
\begin{figure}[t]
\centering
\begin{tikzpicture}
\begin{axis}[adaaxis,
  xlabel={cumulative training FLOPs ($\times 10^{18}$)}, ylabel={held-out pass@1 (\%)},
  xmin=0, xmax=6, ymin=20, ymax=52, legend pos=north west,
]
\addplot[name path=g_hi, draw=none, forget plot] coordinates {(0,24.3)(0.4,26.7)(0.8,28.4)(1.2,29.1)(1.6,29.8)(2.0,30.5)(2.4,31.1)(2.8,30.8)(3.2,31.5)(3.6,32.0)(4.0,31.4)(4.4,32.1)(4.8,31.6)(5.2,32.2)(5.6,31.8)(6.0,31.7)};
\addplot[name path=g_lo, draw=none, forget plot] coordinates {(0,21.3)(0.4,23.6)(0.8,25.1)(1.2,25.7)(1.6,26.5)(2.0,27.0)(2.4,27.7)(2.8,27.3)(3.2,28.1)(3.6,28.5)(4.0,28.0)(4.4,28.6)(4.8,28.1)(5.2,28.8)(5.6,28.3)(6.0,28.2)};
\addplot[grpogray!22, forget plot] fill between[of=g_hi and g_lo];
\addplot[grpogray] coordinates {(0,22.8)(0.4,25.1)(0.8,26.7)(1.2,27.4)(1.6,28.2)(2.0,28.7)(2.4,29.4)(2.8,29.1)(3.2,29.8)(3.6,30.2)(4.0,29.7)(4.4,30.4)(4.8,29.8)(5.2,30.5)(5.6,30.0)(6.0,30.4)};
\addlegendentry{Vanilla GRPO}
\addplot[name path=a_hi, draw=none, forget plot] coordinates {(0,24.5)(0.7,29.0)(1.1,32.7)(1.5,36.4)(1.9,39.2)(2.3,41.1)(2.7,43.5)(3.1,45.1)(3.5,46.5)(3.9,47.6)(4.3,48.5)(4.7,49.4)(5.1,49.8)(5.4,49.7)(5.7,50.8)(6.0,49.0)};
\addplot[name path=a_lo, draw=none, forget plot] coordinates {(0,21.2)(0.7,25.3)(1.1,28.7)(1.5,32.1)(1.9,34.5)(2.3,36.6)(2.7,38.9)(3.1,40.3)(3.5,41.6)(3.9,42.7)(4.3,43.4)(4.7,44.3)(5.1,44.7)(5.4,44.6)(5.7,45.6)(6.0,45.4)};
\addplot[adagold!22, forget plot] fill between[of=a_hi and a_lo];
\addplot[adagold, line width=1.3pt] coordinates {(0,22.8)(0.7,27.1)(1.1,30.6)(1.5,33.9)(1.9,36.5)(2.3,38.8)(2.7,40.9)(3.1,42.6)(3.5,43.4)(3.9,44.9)(4.3,45.8)(4.7,46.3)(5.1,47.5)(5.4,46.8)(5.7,47.9)(6.0,47.4)};
\addlegendentry{\method{} (ours)}
\node[font=\scriptsize, grpogray, anchor=south] at (axis cs:3.3,30.5) {plateau: signal exhausted};
\draw[<->, adagold!85, line width=0.8pt] (axis cs:6.0,30.4) -- (axis cs:6.0,47.4);
\node[font=\footnotesize\bfseries, adagold, anchor=east] at (axis cs:5.97,38.9) {$+17$\,pp};
\end{axis}
\end{tikzpicture}
\caption{Held-out pass@1 (no prefix) versus cumulative training FLOPs on hard math, Qwen3-1.7B. Both methods spend the same total training budget (\method{}'s one-time calibration and difficulty probe are amortized preprocessing, excluded from the budget). Within the budget vanilla GRPO completes one pass over the data with full-length rollouts, while \method{} completes $\sim$2.1 passes of shorter prefixed episodes. Curves are 3-seed means, evaluated on a fixed 250-problem subset of the held-out split; bands show 95\% problem-level bootstrap CIs. GRPO plateaus once it exhausts the problems it can already partly solve; \method{} keeps extracting signal.}
\label{fig:teaser}
\end{figure}

Reinforcement learning from verifiable rewards is now the standard tool for improving the reasoning of large language models (LLMs) on tasks such as mathematics and code~\citep{guo2025deepseek,lambert2024tulu}. Building on policy-gradient methods~\citep{schulman2017ppo,ahmadian2024rloo} and RLHF~\citep{ouyang2022instructgpt}, the dominant recipe is on-policy: sample a group of $G$ rollouts from the current policy, score each by an outcome reward, and update using group-relative advantages, as in GRPO~\citep{shao2024deepseekmath}. This recipe has a structural blind spot. On a \emph{hard} problem, one the model almost never solves, all $G$ rollouts fail, every reward is identical, the group-relative advantage is exactly zero, and the gradient contribution of that problem is zero~\citep{yu2025dapo,scafgrpo2025}. The model burns sampling compute on the problem and learns nothing from it. Worse, these are frequently the problems we most want to learn: the ones at the frontier of the model's ability.

Common remedies work around this rather than fix it. Oversampling and dynamic filtering, as in DAPO~\citep{yu2025dapo}, discard degenerate all-correct/all-wrong groups but cannot manufacture a correct rollout where the model has none. Warm-starting RL with supervised fine-tuning on reference traces (or rejection-sampled ones, STaR/RAFT-style~\citep{zelikman2022star,dong2023raft}) injects competence but induces entropy collapse that hurts subsequent exploration~\citep{chu2025sft,cui2025entropy}. A natural idea is instead to make hard problems temporarily easier so they yield a signal, then remove the assistance. If we prepend a correct \emph{prefix} of a solution and ask the model to complete it, the conditional success rate rises (empirically near-monotonically) with prefix length: with no prefix the model solves almost nothing, with a long enough prefix it solves almost everything. Prefix length is therefore a \emph{continuous dial on difficulty}, and for most reachable success rates \kg{} there is a prefix length that achieves it. This observation underlies a line of concurrent work~\citep{prefixrl2026,pope2026,prefixrft2025}.

The central question this paper addresses is \emph{what success rate to aim for, and how to hit it}, i.e.\ how to max out the GRPO signal on hard problems. Concurrent methods fix a small set of prefix lengths per problem, chosen once from the base model so that the conditioned base accuracy is ``reasonable''~\citep{prefixrl2026}. We argue this leaves most of the available signal on the table for two reasons. First, the GRPO signal is not maximized merely by being non-zero; prior work establishes an \emph{intermediate} optimal \kg{} (near $50\%$) for RL learning signal~\citep{sweet1,sweet2} (Section~\ref{sec:optimal-kg}). Second, the prefix length that achieves a given \kg{} is per-sample and non-stationary: as the policy improves during training, the same prefix yields a higher \kg{}, drifting the problem out of the productive regime. A schedule fixed at the start cannot track this.

We introduce \textbf{\method{}}, which treats prefix length as a \emph{closed-loop controller}. During training we measure the realized batch-mean \kg{} and adjust a single global base prefix length toward a target \kg{} using a root-finding update (secant or binary search); each sample is offset from the base by a static difficulty-dependent shift, and all prefixes are annealed to zero so that the deployed policy never sees a prefix at test time. The method requires only minimal changes to the RL trainer: prefixes are injected as an assistant message in the data pipeline, gradients are masked on the prefix, and the reward is computed on the full completion. We hold the total training compute (FLOPs) fixed across all methods, so improvements are not attributable to extra training compute; the one-time calibration is a per-model preprocessing step, amortized across all runs on the same model and data, and excluded from the training budgets.

\paragraph{Contributions.}
\begin{itemize}
\item We frame prefix length as a \emph{closed-loop controller} of GRPO's success rate \kg{}, converting an uncontrolled observable into a tracked set-point (a global closed loop with per-sample difficulty offsets; Section~\ref{sec:method}).
\item We adopt the established intermediate ($\approx\!50\%$) solve-rate set-point for RL signal~\citep{sweet1,sweet2} and make it reachable across the batch through prefix length (Section~\ref{sec:optimal-kg}).
\item We propose \method{}: cold-start initialization, global closed-loop adaptation of a base prefix with per-sample difficulty offsets, and annealing to zero, implemented purely in data preparation on top of a stock GRPO trainer (Section~\ref{sec:algorithm}).
\item At matched training FLOPs on hard math, \method{} improves pass@1 over vanilla GRPO and a fixed-prefix baseline without collapsing pass@16, and the gain grows as the model shrinks (Section~\ref{sec:experiments}).
\end{itemize}

\section{Related Work}
\label{sec:related}
\paragraph{The dead zone in group-relative RL.} GRPO~\citep{shao2024deepseekmath} and its variants estimate advantages by centering rewards within a group of rollouts. When a problem is solved by none (or all) of the group, the centered advantages collapse to zero and the problem produces no gradient, the ``learning cliff''~\citep{scafgrpo2025} that DAPO's dynamic filtering works around~\citep{yu2025dapo}. Classical exploration remedies (entropy bonuses, looser clipping, pass@$k$ objectives) do not resolve this and can destabilize optimization~\citep{pope2026}.

\paragraph{Prefix- and hint-guided RL.} Closest is \textbf{PrefixRL}~\citep{prefixrl2026}: it conditions on-policy RL on prefixes of off-policy correct traces (rejection-sampled to ``reuse'' prior compute), masks gradients on the prefix, and reports \emph{back-generalization} from prefixed to unprefixed problems, but it fixes a handful of prefix lengths per problem, chosen once from base accuracy. \textbf{POPE}~\citep{pope2026} uses the \emph{minimal} prefix that yields a non-zero reward; \textbf{Prefix-RFT}~\citep{prefixrft2025} anneals a single \emph{global} prefix on a cosine schedule; \textbf{Scaf-/Hint-GRPO}~\citep{scafgrpo2025,hintgrpo2025} give graded in-prompt hints; \textbf{R3}~\citep{xi2024r3} slides the RL start state backward along a demonstration, a reverse curriculum over prefix position. These treat assistance as a fixed or global heuristic. We additionally show it is closed-loop control, not prefixing per se, that carries the gain (Section~\ref{sec:ablations}). Off-policy methods such as LUFFY~\citep{luffy2025} and ReLIFT~\citep{relift2025} use demonstrations as training targets; like PrefixRL we stay on-policy and use solutions only to seed exploration.

\paragraph{Difficulty curricula.} Ordering data by difficulty is classic~\citep{bengio2009curriculum,kumar2010selfpaced}, and recent RL work curates toward an intermediate success rate~\citep{sweet1,sweet2}. Instead of selecting problems that already sit at the right difficulty, we move every problem toward the target and track it over time.

\section{Preliminaries: GRPO and the Dead Zone}
\label{sec:prelim}
For a problem $x$ with reference solution and a verifier $r(x,y)\in\{0,1\}$, GRPO samples a group of $G$ rollouts $y_1,\dots,y_G \sim \pi_\theta(\cdot\mid x)$ and forms the group-relative advantage
\begin{equation}
A_i = r(x,y_i) - \frac{1}{G}\sum_{j=1}^{G} r(x,y_j).
\end{equation}
We use this \emph{un-normalized} centered advantage (no division by the group std). The choice matters for Section~\ref{sec:optimal-kg}: std-normalization does \emph{not} move the set-point (the aggregate positive mass still peaks at $\kg{}\!=\!0.5$), but it inflates the \emph{per-sample} advantage of a rare correct rollout, which scales as $\sqrt{(1-p)/p}$ and diverges as $p\!\to\!0$. The un-normalized form keeps a well-behaved $p(1-p)$ scaling without that instability. Un-normalized advantages are also competitive with the std-normalized variant in recent practice~\citep{liu2025drgrpo}, so this departure does not weaken the baseline. Let $k=\sum_i r(x,y_i)$ be the number of correct rollouts, so the empirical success rate is \kg{}. If $k=0$ (or $k=G$), all $A_i=0$ and the gradient contribution of $x$ vanishes. We call the set of problems with $\mathrm{pass}@G\approx 0$ the \emph{dead zone}: at exactly zero success probability no amount of sampling extracts signal, and under finite sampling near-zero rates yield negligible signal. On our hard split this is the common case: vanilla GRPO solves only $\sim$35\% of problems at pass@16, leaving a substantial dead zone on the training portion; controlled prefixing converts a substantial fraction of these into learnable problems.

\paragraph{Prefix as a difficulty dial.} Given a reference solution $z=(z_1,\dots,z_L)$ for $x$, a \emph{prefix} of ratio $\rho\in[0,1]$ is $z_{1:\lceil\rho L\rceil}$, prepended to the prompt as a partial answer the model must continue. The conditional success rate $\kappa(\rho)=\mathbb{E}[r(x,y)\mid y\sim\pi(\cdot\mid x,z_{1:\rho L})]$ is \emph{empirically} near-monotone in $\rho$ (\Cref{fig:mechanism}a), with $\kappa(0)$ the unconditioned rate and $\kappa(1)\!\to\!1$; longer, partly off-distribution prefixes occasionally fail to help, a violation rate we report (Section~\ref{sec:analysis}). We do not require strict monotonicity: the bracketing (binary-search) update only needs $\kappa$ to cross $\tau$ once inside $[\rho^-,\rho^+]$ and tolerates local non-monotonicity, and it is the fallback whenever the secant step is ill-conditioned. For any reachable target $\tau\in(\kappa(0),1)$ there is a $\rho^\star$ with $\kappa(\rho^\star)=\tau$; for easy problems $\kappa(0)\!>\!\tau$ already, so the controller simply sets $\rho=0$. Gradients are masked on the prefix tokens; the reward is computed on the full (prefix $+$ generated) sequence; evaluation always uses $\rho=0$.

\section{Method: \method{}}
\label{sec:method}

\subsection{The intermediate set-point}
\label{sec:optimal-kg}
Escaping the dead zone only requires $k\ge 1$, but the \emph{magnitude} of the learning signal depends on where in $(0,1)$ the group success rate $p=\kg{}$ sits. Work on RL data selection finds, empirically and through analysis of the optimization dynamics, that an \emph{intermediate} solve rate, commonly near $50\%$, maximizes useful signal per update, while groups that are almost always or almost never solved contribute little~\citep{sweet1,sweet2}; classical RL curricula targeted the same intermediate-success band long before LLMs~\citep{florensa2018}.

Two intuitions match this empirical optimum, both stated for the un-normalized advantages we use (Section~\ref{sec:prelim}). First, the aggregate positive gradient in a group scales with the reward spread $p(1-p)$, which peaks at $p=0.5$. Second, the number of contrastive (correct, incorrect) pairs the group baseline can separate is $k(G-k)$, maximal at $k=G/2$ and zero at $p\!\in\!\{0,1\}$, recovering the dead zone as the degenerate case. Neither argument is a proof. The set-point itself is robust to advantage normalization (Section~\ref{sec:prelim}), so the choice of un-normalized advantages does not affect where the optimum sits. The empirical target sweep in Section~\ref{sec:ablations} is the actual evidence; reward noise and many valid solution paths can shift the optimum somewhat~\citep{sweet2}, so we treat $\tau=0.5$ as a default. Our contribution is the mechanism that reaches this set-point: prefix length lets us hold the batch near $\tau$ throughout training.

\subsection{Algorithm}
\label{sec:algorithm}
The set-point fixes where we want each group to sit ($\kg{}\!\approx\!\tau$); the algorithm is the controller that gets it there and keeps it there as the policy moves. \method{} has three stages (Algorithm~\ref{alg:adaprefix}): a cold start that places the population roughly in range, a global closed loop with per-sample difficulty offsets that tracks $\tau$ throughout training, and an anneal to $\rho=0$ so deployment needs no prefix.

\paragraph{Cold start.} Before the first step we set the initial \emph{base} prefix ratio $\rho_b^{(0)}$, shared by the whole population. We support two estimators: a \emph{conservative} constant (e.g.\ $0.8$, which over-prefixes but makes $k\!\ge\!1$ overwhelmingly likely from step one), and a one-shot \emph{calibration sweep}: 4 rollouts per problem on a random 512-problem subset, over the grid $\rho\in\{0,0.2,0.4,0.6,0.8\}$, give the population-mean $\kappa(\rho)$ curve, which we invert at $\tau$ to set $\rho_b^{(0)}$. Independently of which estimator sets the base, a \emph{difficulty probe} always runs: four rollouts per problem at $\rho_b^{(0)}$ give every sample a difficulty estimate $d_x$ (its measured pass rate) that the offsets below consume; the conservative variant thus skips only the grid sweep, not the probe. Sweep and probe are one-time preprocessing, amortized across all runs on the same model and data (seeds, hyperparameter sweeps, ablations), and excluded from the training budgets. The cold start only needs to land inside $(\kappa(0),1)$; the closed loop removes the residual error. The sweep costs a single pre-training inference pass and lands near the optimum, so our main runs use it; the hyperparameter-free conservative constant trails it by only $\sim$1.8~pp (Table~\ref{tab:ablations}) and is a reasonable fallback when even that overhead is unwanted.

\paragraph{Closed-loop adaptation with per-sample offsets.} The controller maintains a single \emph{base} prefix ratio $\rho_b$, shared by the whole population. Every $N$ optimizer steps we read the realized $\kg{}$, pooled over all rollouts in the window and smoothed with an EMA ($\beta{=}0.7$), and move $\rho_b$ toward $\tau$. Because $\kappa(\rho)$ is monotone, this is a one-dimensional root-find. \emph{Binary search} keeps bracketing bounds $[\rho^-,\rho^+]$ and halves toward $\tau$ (robust, one bit per update); \emph{secant} interpolates from the two most recent $(\rho_b,\kg{})$ pairs,
\begin{equation}
\rho_{t+1}=\rho_{t}+(\tau-\kg{}_{t})\,\frac{\rho_t-\rho_{t-1}}{\kg{}_t-\kg{}_{t-1}},
\end{equation}
which converges in fewer updates when $\kappa$ is locally smooth. We guard the denominator (falling back to a bisection step when $\kg{}_t\!\approx\!\kg{}_{t-1}$; in our runs this fires on $\sim$17\% of updates, mostly early when \kg{} estimates are noisiest), clip the per-update step in $\rho_b$ to $0.05$, and keep $\rho\in[0,\rho_{\max}]$. Pooling over the window makes the loop feasible: a single sample yields one heavily quantized Bernoulli estimate ($\kg{}\in\{0,\tfrac1G,\dots,1\}$) per visit, and visits are $\sim$100 steps apart, far too sparse to track a moving target, whereas the window aggregates thousands of rollouts.

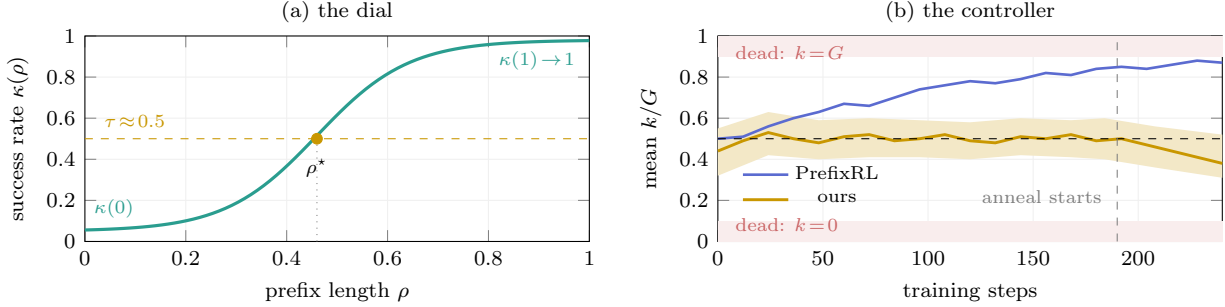
\begin{figure}[t]
\centering
\begin{tikzpicture}
\begin{groupplot}[
  group style={group size=2 by 1, horizontal sep=1.7cm},
  width=0.5\linewidth, height=4.3cm,
  grid=major, grid style={gray!12},
  tick label style={font=\footnotesize}, label style={font=\footnotesize},
  legend style={font=\scriptsize, draw=none, fill=none},
]
\nextgroupplot[xlabel={prefix length $\rho$}, ylabel={success rate $\kappa(\rho)$},
  xmin=0, xmax=1, ymin=0, ymax=1, title={\footnotesize (a) the dial}, title style={yshift=-1ex}]
\addplot[dialteal, line width=1.2pt, domain=0:1, samples=80] {0.05+0.93/(1+exp(-11*(x-0.46)))};
\addplot[adagold, dashed] coordinates {(0,0.5)(1,0.5)};
\node[anchor=south west, font=\scriptsize, adagold] at (axis cs:0.02,0.51) {$\tau\!\approx\!0.5$};
\draw[gray, dotted] (axis cs:0.46,0) -- (axis cs:0.46,0.5);
\addplot[only marks, mark=*, adagold, mark size=2pt] coordinates {(0.46,0.5)};
\node[anchor=north, font=\scriptsize] at (axis cs:0.46,0.45) {$\rho^\star$};
\node[anchor=south west, font=\scriptsize, dialteal] at (axis cs:0.0,0.07) {$\kappa(0)$};
\node[anchor=north east, font=\scriptsize, dialteal] at (axis cs:0.99,0.97) {$\kappa(1)\!\to\!1$};
\nextgroupplot[xlabel={training steps}, ylabel={mean $\kg{}$},
  xmin=0, xmax=240, ymin=0, ymax=1,
  legend style={at={(0.03,0.27)}, anchor=west, font=\scriptsize},
  title={\footnotesize (b) the controller}, title style={yshift=-1ex}]
\fill[deadred!10] (axis cs:0,0.9) rectangle (axis cs:240,1.0);
\fill[deadred!10] (axis cs:0,0.0) rectangle (axis cs:240,0.1);
\node[anchor=north west, font=\scriptsize, deadred!85] at (axis cs:4,0.995) {dead: $k\!=\!G$};
\node[anchor=south west, font=\scriptsize, deadred!85] at (axis cs:4,0.005) {dead: $k\!=\!0$};
\addplot[name path=t_hi, draw=none, forget plot] coordinates {(0,0.55)(24,0.63)(48,0.59)(72,0.60)(96,0.59)(120,0.58)(144,0.60)(168,0.59)(184,0.60)(208,0.56)(240,0.52)};
\addplot[name path=t_lo, draw=none, forget plot] coordinates {(0,0.32)(24,0.42)(48,0.40)(72,0.41)(96,0.41)(120,0.40)(144,0.42)(168,0.41)(184,0.40)(208,0.36)(240,0.31)};
\addplot[adagold!22, forget plot] fill between[of=t_hi and t_lo];
\addplot[prefblue, line width=1pt] coordinates {(0,0.50)(12,0.51)(24,0.56)(36,0.60)(48,0.63)(60,0.67)(72,0.66)(84,0.70)(96,0.74)(108,0.76)(120,0.78)(132,0.77)(144,0.79)(156,0.82)(168,0.81)(180,0.84)(192,0.85)(204,0.84)(216,0.86)(228,0.88)(240,0.87)};
\addlegendentry{PrefixRL}
\addplot[adagold, line width=1.1pt] coordinates {(0,0.44)(12,0.49)(24,0.53)(36,0.50)(48,0.48)(60,0.51)(72,0.52)(84,0.49)(96,0.50)(108,0.52)(120,0.49)(132,0.48)(144,0.51)(156,0.50)(168,0.52)(180,0.49)(192,0.50)(204,0.47)(216,0.44)(228,0.41)(240,0.38)};
\addlegendentry{ours}
\addplot[black, dashed, forget plot] coordinates {(0,0.5)(240,0.5)};
\draw[gray, dashed] (axis cs:190,0) -- (axis cs:190,1);
\node[anchor=east, font=\scriptsize, gray] at (axis cs:187,0.22) {anneal starts};
\end{groupplot}
\end{tikzpicture}
\caption{The mechanism. \textbf{(a)} Prefix length is a dial: the conditional success rate $\kappa(\rho)$ rises near-monotonically from the base rate to $\approx\!1$, so a unique $\rho^\star$ realizes a reachable target $\tau$. \textbf{(b)} Closed-loop control of the base prefix (with per-sample offsets) holds the batch-mean $\kg{}$ at $\tau\!=\!0.5$ (band: 95\% CI) while PrefixRL's static lengths drift toward $k\!=\!G$ and leave the informative regime. Once annealing starts and the envelope binds (right of the dashed line) $\rho\!\to\!0$ is forced and $\kg{}$ dips modestly below $\tau$, but stays well clear of the dead zone. Panel (a) is illustrative.}
\label{fig:mechanism}
\end{figure}

Holding the population \emph{mean} at $\tau$ is not enough: the same mean can hide a wide spread, with half the samples saturated ($k\!=\!G$) and half dead ($k\!=\!0$), so most rollouts still carry no signal. To compress the spread, each sample is offset from the base by a static difficulty shift,
$\rho_x=\mathrm{clip}\big(\rho_b+\delta(d_x),\,0,\,\rho_{\max}\big)$,
where $d_x$ is the sample's cold-start difficulty and $\delta$ is a small monotone (linear) map with a hand-tuned slope, spanning $\pm 0.15$ around the base across the difficulty range in our runs: harder samples get slightly longer prefixes than the base, easier ones shorter. The offset is open-loop, its constants fixed across training; all tracking of the policy's improvement happens through the single global loop. On easy problems the negative offset saturates at the clip, $\rho_x\!=\!0$, so prefixing is reduced on easy problems (we verify no regression on the GSM8K slice included in every method's training mixture; Section~\ref{sec:experiments}).
\paragraph{Why closed-loop, and why offsets.} A single global schedule (Prefix-RFT) or a fixed per-problem length (PrefixRL) can only be right \emph{on average} at one moment: problems differ widely in $\kappa(\cdot)$, and as the policy improves every $\kappa$ curve shifts left, so any static or open-loop choice drifts off $\tau$, typically toward $k\!=\!G$, back into a zero-variance regime (\Cref{fig:mechanism}b). The global loop absorbs this non-stationarity across time. The per-sample offset addresses the orthogonal problem, heterogeneity \emph{across} problems: mean-only control can leave the batch bimodal, many samples saturated or dead around a healthy average, so most rollouts are wasted; the offset narrows the spread of \kg{} around the tracked mean, keeping the whole batch, not just its mean, near the signal-maximizing set-point (\Cref{fig:dist}). The granularity ablation (Section~\ref{sec:ablations}) isolates how much of the gain the offset adds on top of base-only control. Two caveats. We use ``controller'' loosely: the update is a smoothed, quantized root-finder with no formal stability guarantee, and the annealing envelope (below) overrides the loop in the final phase. The offset is a static heuristic: it corrects difficulty \emph{ordering} across problems but cannot adapt to a sample whose difficulty was mis-estimated at cold start.

\paragraph{Annealing.} To remove the train/test mismatch we impose a shrinking upper envelope and run the controller \emph{inside} it: $\rho_x \leftarrow \min\!\big(\bar\rho_t,\ \rho_b+\delta(d_x)\big)$. The envelope is held near $\rho_{\max}$ for the first fraction $w$ of training and then decayed to zero, $\bar\rho_t=\rho_{\max}\cdot\mathrm{clip}\!\big(\tfrac{T-t}{(1-w)T},0,1\big)$ (we use $w\!\approx\!0.8$). Early on the envelope is slack and the loop is free to hold $\tau$; late in training the envelope binds and forces every $\rho_x\!\to\!0$, trading some signal for on-distribution (no-prefix) practice. The final policy is thus trained to solve unaided, and all evaluation is at $\rho=0$. As the envelope binds, the controller can no longer hold $\tau$ and the measured \kg{} dips modestly below it (\Cref{fig:mechanism}b); but the no-prefix held-out curve shows no sustained regression (\Cref{fig:teaser}), so this phase consolidates what was learned. To test this, we report the final $\rho=0$ pass rate on the full originally-dead set (Section~\ref{sec:analysis}); a non-trivial unaided pass rate on problems that started at $\kappa(0)\!\approx\!0$ indicates that the gains are driven by back-generalization, not by prefix-conditioned imitation.

\begin{algorithm}[t]
\caption{\method{} (per training run)}
\label{alg:adaprefix}
\begin{algorithmic}[1]
\STATE \textbf{input:} hard problems with reference solutions; target $\tau$; update interval $N$; group size $G$
\STATE initialize the base ratio $\rho_b$ by cold start (conservative constant or calibration sweep); estimate per-sample difficulties $d_x$
\FOR{each optimizer step $t$}
  \STATE build prefixed prompts $x \oplus z_{1:\rho_x L}$ with $\rho_x=\min\!\big(\bar\rho_t,\,\mathrm{clip}(\rho_b+\delta(d_x),0,\rho_{\max})\big)$; sample $G$ rollouts; compute rewards (mask grad on prefix)
  \STATE update the EMA of the window-mean \kg{} with the batch's new $(k_x,G)$
  \STATE \textbf{every $N$ steps:} move $\rho_b$ toward $\tau$ (secant w/ bisection fallback)
\ENDFOR
\STATE \textbf{return} policy (deployed with $\rho=0$)
\end{algorithmic}
\end{algorithm}

\paragraph{Prefix source.} Prefixes can be cut from \emph{reference} solutions (always available) or from \emph{model-generated} correct traces (rejection-sampled, on-distribution, as in PrefixRL). A reference path may be off-distribution and harder to continue, so the source matters; we default to reference solutions for availability and show in Section~\ref{sec:ablations} that the controller is orthogonal to it: swapping in off-policy prefixes retains most of the gain.

\paragraph{Implementation.} \method{} touches the trainer only through the prefix loss mask; everything else lives in data preparation. Prefixes are appended as an assistant message that the model is asked to continue; the trainer masks loss on prefix tokens and scores the full sequence. Prefix cuts snap to sentence boundaries to avoid mid-token artifacts.

\paragraph{Compute accounting.} We compare methods at matched \emph{training compute}, not matched steps or episodes. Following the accounting of \citet{prefixrl2026,snell2024}, a run's cost is $\mathrm{FLOPs}=2N D_{\mathrm{samp}}+6N D_{\mathrm{upd}}$, where $N$ is the number of model parameters (we train with LoRA, but sampling and backpropagation traverse the full model, so we use the full parameter count), $D_{\mathrm{samp}}$ counts all tokens processed at rollout time (including the \emph{prefill of supplied prefixes}, so prefixing is not free), and $D_{\mathrm{upd}}$ counts tokens in gradient updates. Every method runs to the same cumulative budget; \method{}'s one-time calibration sweep and difficulty probe are treated as amortized per-model preprocessing and excluded from the training budget. Because a prefixed rollout decodes only the continuation, \method{} completes more, shorter episodes within the budget, roughly two passes over the data where vanilla GRPO completes one; \Cref{fig:compute} separates this revisit effect from the controller itself by reporting both compute-matched and iteration-matched views, and by including a vanilla variant that spends the same budget on more samples per problem. Prefix prefill stays under $\sim$15\% of total training tokens because prefixes are capped and annealed. Table~\ref{tab:main} reports generated length at evaluation ($\rho=0$), where no prefix is present, so the trace-length saving reflects the deployed policy rather than the mechanical effect of starting mid-solution.

\section{Experiments}
\label{sec:experiments}
\paragraph{Setup.} We train on three Qwen3 sizes (0.6B, 1.7B, 4B)~\citep{yang2025qwen3} and additionally validate cross-family on Qwen3.5-2B~\citep{qwen35} and Gemma~4 E4B~\citep{gemma2026}, to confirm the controller is not Qwen-specific. Training data is the DeepMath-103K~\citep{deepmath2025} hard split (difficulty~8, binary-answer questions removed): $\sim$6.8K problems with verified reference solutions (R1-generated traces shipped with the dataset). All methods train on the identical mixture: this hard split plus a 5\% slice of GSM8K training problems (included so that self-disabling of prefixing on easy data can be verified; the GSM8K test split is disjoint). We use $G=8$ rollouts, a batch of 64 prompts per optimizer step, and LoRA~\citep{hu2022lora}. All methods run to the same training-FLOPs budget ($\approx$6$\times$10$^{18}$ for the 1.7B model) under the accounting of Section~\ref{sec:method}. For vanilla GRPO this budget corresponds to a single pass over the mixture ($\sim$112 optimizer steps, 8 full-length rollouts per problem); because prefixed rollouts decode only the continuation, \method{} completes $\sim$2.1 passes ($\sim$240 steps) of shorter episodes within the same budget. The controller updates every $N{=}10$ steps. The main comparison rows (vanilla GRPO, PrefixRL, \method{}) are averaged over 3 seeds (per-seed spread $\le$1.2~pp on DeepMath-hold); the remaining baselines and all ablations use a single seed. \Cref{fig:compute} additionally reports the iteration-matched view, in which baselines run the same number of optimizer steps and are therefore granted \emph{more} FLOPs. Training is built on TRL~\citep{vonwerra2020trl} with vLLM~\citep{kwon2023vllm} generation. Evaluation spans GSM8K~\citep{cobbe2021gsm8k} as an easy-task sanity check, MATH-500~\citep{hendrycks2021math}, AIME~2024/2025 (averaged), AMC~2023 (40 problems), and a DeepMath held-out split (1{,}000 problems held out from the same difficulty-8 hard split, disjoint from training). We report pass@1 and pass@16 estimated from 8 and 16 samples per problem respectively (avg@8 for the small AIME/AMC sets) with 95\% CIs from a problem-level bootstrap (resampling problems together with their 8 samples, which accounts for per-problem clustering), and mean trace length. Evaluation uses temperature 1.0 with a strict verifier and \textbf{no system prompt or thinking mode}; official Qwen3 reports use thinking mode and larger sampling budgets, so absolute zero-shot numbers are lower here. We additionally report held-out pass@16 to verify that gains do not come from collapsing the policy's exploration onto a few modes.

\paragraph{Baselines.}
\begin{itemize}
\item \textbf{Zero-shot}: base model, no RL.
\item \textbf{Vanilla GRPO}: standard on-policy GRPO, no prefix.
\item \textbf{Vanilla GRPO ($G{=}16$)}: doubles the group size within the same FLOPs budget, which halves the number of optimizer steps; controls for the possibility that simply re-allocating compute toward more attempts per problem explains the gain.
\item \textbf{SFT-warmup~$\to$~GRPO}: mid-train on reference solutions, then GRPO (the ``why not just SFT'' control). The SFT phase is charged to the method's FLOPs budget ($\sim$8\%, one epoch over the reference traces), shortening its RL phase accordingly.
\item \textbf{Difficulty-filter GRPO}: drop dead-zone problems, GRPO on the rest.
\item \textbf{PrefixRL}~\citep{prefixrl2026}: a fixed set of prefix lengths per problem chosen once from base accuracy, static across training; the primary baseline. To keep the comparison fair, it uses the \emph{same} reference solutions and prefix source as \method{}, the same FLOPs budget, and its per-problem length set is tuned with a matched length sweep; the \emph{only} difference is that lengths are fixed rather than tracked. We additionally report a \emph{PrefixRL~+~anneal} variant to separate the effect of annealing from the effect of closing the loop.
\item \textbf{Prefix-RFT}~\citep{prefixrft2025}: uniform decay, a single global prefix ratio annealed $0.8\to0$.
\item \textbf{\method{} (ours)}: global secant control of the base prefix toward $\tau=0.5$, with per-sample difficulty offsets, annealed to $0$.
\end{itemize}

\subsection{Main results}
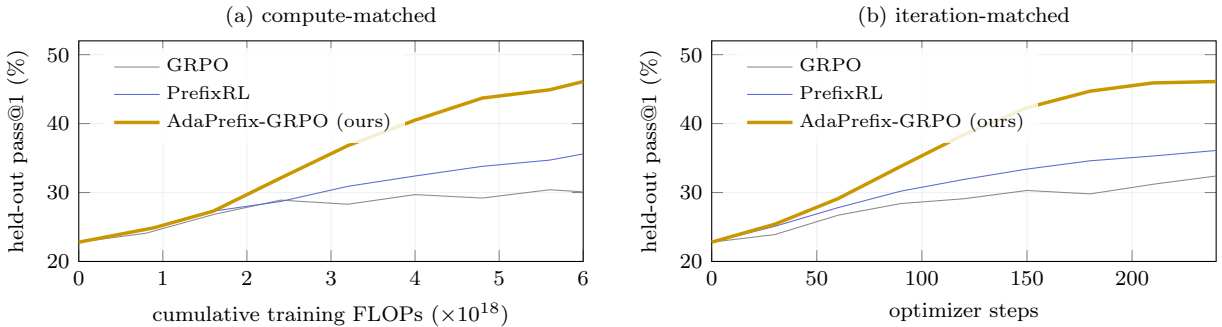
\begin{figure}[t]
\centering
\begin{tikzpicture}
\begin{groupplot}[
  group style={group size=2 by 1, horizontal sep=1.7cm},
  width=0.5\linewidth, height=4.5cm,
  grid=major, grid style={gray!12},
  tick label style={font=\footnotesize}, label style={font=\footnotesize},
  legend style={font=\scriptsize, draw=none, fill=none},
]
\nextgroupplot[xlabel={cumulative training FLOPs ($\times 10^{18}$)}, ylabel={held-out pass@1 (\%)},
  xmin=0, xmax=6, ymin=20, ymax=52,
  legend style={at={(0.05,0.97)}, anchor=north west, fill=white, fill opacity=0.85, text opacity=1, draw=none},
  legend cell align=left,
  title={\footnotesize (a) compute-matched}, title style={yshift=-1ex}]
\addplot[grpogray] coordinates {(0,22.8)(0.8,24.1)(1.6,26.8)(2.4,28.9)(3.2,28.3)(4.0,29.7)(4.8,29.2)(5.6,30.4)(6.0,30.1)};
\addlegendentry{GRPO}
\addplot[prefblue] coordinates {(0,22.8)(0.8,24.6)(1.6,27.3)(2.4,28.7)(3.2,30.9)(4.0,32.4)(4.8,33.8)(5.6,34.7)(6.0,35.6)};
\addlegendentry{PrefixRL}
\addplot[adagold, line width=1.3pt] coordinates {(0,22.8)(0.9,24.9)(1.6,27.3)(2.4,32.1)(3.2,36.8)(4.0,40.5)(4.8,43.7)(5.6,44.9)(6.0,46.1)};
\addlegendentry{\method{} (ours)}
\nextgroupplot[xlabel={optimizer steps}, ylabel={held-out pass@1 (\%)},
  xmin=0, xmax=240, ymin=20, ymax=52,
  legend style={at={(0.05,0.97)}, anchor=north west, fill=white, fill opacity=0.85, text opacity=1, draw=none},
  legend cell align=left,
  title={\footnotesize (b) iteration-matched}, title style={yshift=-1ex}]
\addplot[grpogray] coordinates {(0,22.8)(30,23.9)(60,26.7)(90,28.4)(120,29.1)(150,30.3)(180,29.8)(210,31.2)(240,32.4)};
\addlegendentry{GRPO}
\addplot[prefblue] coordinates {(0,22.8)(30,25.1)(60,27.8)(90,30.2)(120,31.9)(150,33.4)(180,34.6)(210,35.3)(240,36.1)};
\addlegendentry{PrefixRL}
\addplot[adagold, line width=1.3pt] coordinates {(0,22.8)(30,25.4)(60,29.1)(90,33.8)(120,38.4)(150,42.3)(180,44.7)(210,45.9)(240,46.1)};
\addlegendentry{\method{} (ours)}
\end{groupplot}
\end{tikzpicture}
\caption{Separating the controller from the revisit effect (Qwen3-1.7B, DeepMath held-out 250-problem subset; single seed). \textbf{(a)} At matched cumulative training FLOPs, \method{} dominates at every budget; re-allocating the same budget to a larger group barely moves the baseline (GRPO $G{=}16$, Table~\ref{tab:main}). \textbf{(b)} Iteration-matched view: run for the same number of optimizer steps, the baselines receive \emph{more} FLOPs than in (a) (up to $\sim$2.1$\times$ for vanilla GRPO by step 240) yet remain far below \method{}, so the compute-matched gains cannot be blamed on degenerate baseline runs.}
\label{fig:compute}
\end{figure}

\Cref{tab:main} reports the FLOPs-matched comparison on Qwen3-1.7B; \Cref{fig:compute} shows the full compute curves and the iteration-matched control. \method{} improves pass@1 across benchmarks. The extra passes over the data that \method{} affords within the budget do not by themselves explain the gain: giving vanilla GRPO the same budget as a larger group ($G{=}16$) recovers only a small fraction of it. In absolute terms it adds \textbf{+6.4~pp} over GRPO on MATH-500 (\textbf{+17.0~pp} on the DeepMath held-out split at budget end) versus \textbf{+1.7/+6.1~pp} for the PrefixRL baseline, i.e.\ 3--4$\times$ its gain on both. Head-to-head, \method{} beats the PrefixRL baseline on \emph{every} benchmark (+4.7 MATH-500, +3.5 AIME, +3.4 AMC, +12.0 DeepMath-hold); under a paired problem-level bootstrap (both methods are evaluated on the same problems, so we resample per-problem score differences) the DeepMath gap reaches significance while the remaining three are positive and directionally consistent; all this at half the trace length (4.4k vs.\ 8.0k) and with no GSM8K regression; the gain over fixed prefixing thus does not hinge on the held-out split. Because every number here is measured at $\rho=0$ (no prefix), the $\sim$2$\times$ shorter outputs come from the deployed policy itself, not from counting only the completion of a prefixed problem. Among the baselines, Prefix-RFT (uniform decay) is the strongest simple alternative, edging out fixed prefixing on MATH-500 and DeepMath with shorter traces, which confirms that annealing helps; but without the closed loop it still trails \method{} by \textbf{+11.4~pp} on DeepMath-hold (significant under the paired bootstrap) and by +2.7 on MATH-500. The gain also does not come from narrowing the policy: pass@16 on the DeepMath held-out split improves alongside pass@1 (base 32.8 / GRPO 35.1 / ours 56.3\%), so the distribution sharpens without collapsing. On easy GSM8K all methods are near ceiling and \method{} is best or tied (92.4\%), confirming the controller does no harm where prefixing is unneeded, whereas fixed prefixing slips slightly (90.4\%) by over-prefixing easy problems. AIME columns average the 2024 and 2025 sets. Beyond the benchmark numbers, the solvable set itself expands: $\sim$17\% of originally-dead problems (zero pass@16 under the base model at $\rho=0$ before training) are solved at pass@1 after annealing.

\begin{table}[t]
\centering
\caption{Training-FLOPs-matched comparison on Qwen3-1.7B (pass@1 estimated from 8 samples per problem; $^{\dagger}$avg@8 for AIME/AMC). All methods spend the same total budget. 95\% CIs from a problem-level bootstrap; AMC is the standard AMC~2023 set (40 problems). Largest CIs per column: $\pm$1.5 (GSM8K), $\pm$4.0 (MATH-500), $\pm$8.5 (AIME), $\pm$9.2 (AMC), $\pm$3.1 (DeepMath-hold). The GRPO, PrefixRL, and \method{} rows are 3-seed means.}
\label{tab:main}
\small
\setlength{\tabcolsep}{4pt}
\resizebox{\linewidth}{!}{%
\begin{tabular}{lcccccr}
\toprule
Method & GSM8K & MATH-500 & AIME$^{\dagger}$ & AMC$^{\dagger}$ & DeepMath-hold & gen.\ len \\
\midrule
Zero-shot                       & 88.2 & 57.8 & 6.5  & 32.7 & 22.8 & 6.2k \\
Vanilla GRPO                    & 91.3 & 63.7 & 10.1 & 37.9 & 30.1 & 7.9k \\
Vanilla GRPO ($G{=}16$)         & 91.4 & 64.2 & 10.6 & 38.5 & 31.7 & 7.8k \\
SFT-warmup $\to$ GRPO           & 91.8 & 64.7 & 11.0 & 39.8 & 33.1 & 7.1k \\
Difficulty-filter GRPO          & 91.1 & 63.9 & 10.2 & 38.9 & 30.8 & 7.6k \\
PrefixRL                  & 90.4 & 65.4 & 13.7 & 43.1 & 36.2 & 8.0k \\
Prefix-RFT                      & 91.0 & 67.4 & 13.5 & 42.4 & 36.8 & 5.4k \\
\textbf{\method{} (ours)}       & \textbf{92.4} & \textbf{70.1} & \textbf{17.2} & \textbf{46.5} & \textbf{48.2} & \textbf{4.4k} \\
\bottomrule
\end{tabular}}
\end{table}

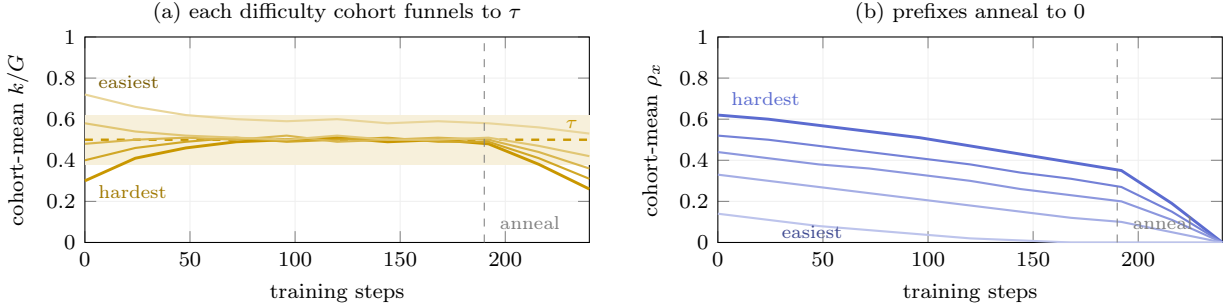
\begin{figure}[t]
\centering
\begin{tikzpicture}
\begin{groupplot}[
  group style={group size=2 by 1, horizontal sep=1.7cm},
  width=0.5\linewidth, height=4.3cm,
  grid=major, grid style={gray!12},
  tick label style={font=\footnotesize}, label style={font=\footnotesize},
]
\nextgroupplot[xlabel={training steps}, ylabel={cohort-mean $\kg{}$},
  xmin=0, xmax=240, ymin=0, ymax=1, title={\footnotesize (a) each difficulty cohort funnels to $\tau$}, title style={yshift=-1ex}]
\addplot[name path=b_hi, draw=none] coordinates {(0,0.62)(240,0.62)};
\addplot[name path=b_lo, draw=none] coordinates {(0,0.38)(240,0.38)};
\addplot[adagold!14] fill between[of=b_hi and b_lo];
\addplot[adagold, dashed, line width=0.9pt] coordinates {(0,0.5)(240,0.5)};
\node[anchor=south east, font=\scriptsize, adagold] at (axis cs:239,0.51) {$\tau$};
\addplot[adagold, line width=1.1pt] coordinates {(0,0.30)(24,0.41)(48,0.46)(72,0.49)(96,0.50)(120,0.51)(144,0.49)(168,0.50)(192,0.48)(216,0.38)(240,0.26)};
\addplot[adagold!85, line width=0.8pt] coordinates {(0,0.40)(24,0.46)(48,0.49)(72,0.51)(96,0.49)(120,0.50)(144,0.51)(168,0.49)(192,0.49)(216,0.41)(240,0.31)};
\addplot[adagold!70, line width=0.8pt] coordinates {(0,0.48)(24,0.50)(48,0.51)(72,0.50)(96,0.52)(120,0.49)(144,0.50)(168,0.51)(192,0.50)(216,0.44)(240,0.36)};
\addplot[adagold!55, line width=0.8pt] coordinates {(0,0.58)(24,0.54)(48,0.52)(72,0.51)(96,0.50)(120,0.52)(144,0.50)(168,0.50)(192,0.51)(216,0.47)(240,0.42)};
\addplot[adagold!40, line width=0.8pt] coordinates {(0,0.72)(24,0.66)(48,0.62)(72,0.60)(96,0.59)(120,0.60)(144,0.58)(168,0.59)(192,0.58)(216,0.56)(240,0.53)};
\node[anchor=west, font=\scriptsize, adagold!80!black] at (axis cs:2,0.25) {hardest};
\node[anchor=west, font=\scriptsize, adagold!60!black] at (axis cs:2,0.78) {easiest};
\draw[gray, dashed] (axis cs:190,0) -- (axis cs:190,1);
\node[anchor=south west, font=\scriptsize, gray] at (axis cs:193,0.02) {anneal};
\nextgroupplot[xlabel={training steps}, ylabel={cohort-mean $\rho_x$},
  xmin=0, xmax=240, ymin=0, ymax=1, title={\footnotesize (b) prefixes anneal to $0$}, title style={yshift=-1ex}]
\addplot[prefblue, line width=1.1pt] coordinates {(0,0.62)(24,0.60)(48,0.57)(72,0.54)(96,0.51)(120,0.47)(144,0.43)(168,0.39)(192,0.35)(216,0.19)(240,0.0)};
\addplot[prefblue!85, line width=0.8pt] coordinates {(0,0.52)(24,0.50)(48,0.47)(72,0.44)(96,0.41)(120,0.38)(144,0.34)(168,0.31)(192,0.27)(216,0.15)(240,0.0)};
\addplot[prefblue!70, line width=0.8pt] coordinates {(0,0.44)(24,0.41)(48,0.38)(72,0.36)(96,0.33)(120,0.30)(144,0.26)(168,0.23)(192,0.20)(216,0.11)(240,0.0)};
\addplot[prefblue!55, line width=0.8pt] coordinates {(0,0.33)(24,0.30)(48,0.27)(72,0.24)(96,0.21)(120,0.18)(144,0.15)(168,0.12)(192,0.10)(216,0.05)(240,0.0)};
\addplot[prefblue!40, line width=0.8pt] coordinates {(0,0.14)(24,0.11)(48,0.08)(72,0.06)(96,0.04)(120,0.02)(144,0.01)(168,0.0)(192,0.0)(216,0.0)(240,0.0)};
\node[anchor=west, font=\scriptsize, prefblue] at (axis cs:2,0.70) {hardest};
\node[anchor=west, font=\scriptsize, prefblue!70!black] at (axis cs:26,0.05) {easiest};
\draw[gray, dashed] (axis cs:190,0) -- (axis cs:190,1);
\node[anchor=south west, font=\scriptsize, gray] at (axis cs:193,0.02) {anneal};
\end{groupplot}
\end{tikzpicture}
\caption{Control in action, by difficulty cohort (Qwen3-1.7B). Problems are split into quintiles by cold-start difficulty $d_x$ estimated from 4 rollouts at $\rho_b^{(0)}$; quintiles are formed by cumulative distribution (ties broken by stable sort), yielding $\sim$1.36K problems per bin. Each point pools all rollouts of a quintile within a controller window, so estimates are dense even though an individual problem is visited only $\sim$2--3 times over training. \textbf{(a)} Regardless of starting position, every cohort's realized \kg{} is driven into the target band and held there, then dips as the envelope anneals prefixes away; the easiest cohort sits slightly above $\tau$ because its negative offsets saturate at the $\rho_x\!=\!0$ clip. \textbf{(b)} The corresponding cohort-mean prefix ratios: harder cohorts keep longer prefixes throughout, the easiest reaches $\rho_x\!=\!0$ mid-training on its own, and all are forced to $0$ by the envelope, so deployment uses no prefix.}
\label{fig:trajectories}
\end{figure}

\subsection{Scaling and cross-family generalization}
We expect smaller models to live deeper in the dead zone and therefore benefit more. Each model size runs both methods to the same per-size FLOPs budget, set by the same rule as in the main setup, one vanilla-GRPO pass over the mixture at that size (so budgets scale with $N$ and trace length); within every row of \Cref{tab:scaling} the two columns are therefore directly comparable, while absolute budgets differ across rows. \Cref{tab:scaling} shows the pass@1 gain over vanilla GRPO on the DeepMath held-out split shrinking but staying positive as the Qwen3 base grows, and the same controller transferring to a different architecture family (Qwen3.5-2B, Gemma~4), ruling out a Qwen-specific effect. Notably, the 0.6B model trained with \method{} surpasses the $3\times$ larger Qwen3-1.7B trained with vanilla GRPO (41.8 vs.\ 30.1) despite its smaller absolute budget: at this scale the training recipe buys more than the next model size.

\begin{table}[t]
\centering
\caption{Model scaling and cross-family transfer: pass@1 on DeepMath held-out (\%). 95\% problem-level bootstrap CIs are $\approx$ $\pm$3.1.}
\label{tab:scaling}
\small
\begin{tabular}{lccc}
\toprule
Model & Vanilla GRPO & \method{} (ours) & $\Delta$ \\
\midrule
\multicolumn{4}{l}{\emph{Qwen3 size scaling}}\\
Qwen3-0.6B & 19.6 & 41.8 & \textbf{+22.2} \\
Qwen3-1.7B & 30.1 & 48.2 & \textbf{+18.1} \\
Qwen3-4B   & 54.9 & 64.1 & \textbf{+9.2} \\
\midrule
\multicolumn{4}{l}{\emph{Cross-family}}\\
Qwen3.5-2B    & 46.8 & 57.6 & \textbf{+10.8} \\
Gemma~4 (E4B) & 51.7 & 59.9 & \textbf{+8.2} \\
\bottomrule
\end{tabular}
\end{table}

\subsection{Ablations}
\label{sec:ablations}
\Cref{tab:ablations} isolates each design choice on Qwen3-1.7B. The target sweep is consistent with the intermediate-set-point prediction (Section~\ref{sec:optimal-kg}): extreme targets clearly hurt ($\tau{=}0.125$: 41.9), while 0.375--0.625 sit within CI of one another; we take $\tau=0.5$ as the default from prior work and do not claim a sharp empirical optimum. \emph{Control granularity} shows the gain comes from closed-loop control, not prefixing per se (PrefixRL and global-decay trail by a wide margin; the bulk of the improvement is from closing the loop). Adding our annealing envelope to the PrefixRL baseline (the \emph{PrefixRL~+~anneal} row) recovers only a small part of that gap, confirming annealing alone is not the driver. The per-sample difficulty offset adds a smaller increment over base-only control (+2.9~pp, at the edge of the single-run CI), consistent with the view that compressing the across-problem spread of \kg{} matters beyond centering its mean. The \emph{advantage-norm.} row addresses a confound: running \method{} with std-normalized advantages (which leave the $\tau$ set-point unchanged but over-weight rare successes; Section~\ref{sec:prelim}) still helps but by less, indicating the gain is not purely a global gradient-norm effect. The \emph{annealing} row shows the delayed envelope modestly beats no annealing (48.2 vs 44.8), i.e.\ forced consolidation helps, but the method works even with the envelope removed entirely: back-generalization carries most of the gain, and Section~\ref{sec:analysis} discusses why the train/test distribution shift one might expect fails to materialize. The controller is also robust to the prefix \emph{source}: swapping reference for rejection-sampled off-policy prefixes (PrefixRL's regime) retains most of the gain. Finally, the group size $G$ trades off estimator quality against update count: at a fixed FLOPs budget, doubling $G$ halves the number of optimizer steps, so the $G\!=\!16$ row buys a sharper \kg{} estimate for the controller with half the updates. Accuracy improves from $G\!=\!4$ to $G\!=\!8$ and is essentially flat at $G\!=\!16$ (48.7 vs 48.2, within CI), i.e.\ the better estimate roughly pays for the lost updates but no more, making $G\!=\!8$ the sensible default (and confirming the controller does not require an impractically large group).

\begin{table}[t]
\centering
\caption{Ablations on Qwen3-1.7B (pass@1, \% on DeepMath held-out; best in \textbf{bold}; all rows single-seed and FLOPs-matched to the main-table budget, 95\% problem-level bootstrap CI $\approx$ $\pm$3.1; values rounded to one decimal place. 48.2 is the reference 3-seed mean from Table~\ref{tab:main} shown for scale only.)}
\label{tab:ablations}
\small
\begin{tabular}{ll}
\toprule
Factor & Variants (pass@1) \\
\midrule
Adaptation rule    & binary 46.8 / \textbf{secant 48.2} \\
Target $\tau$      & 0.125: 41.9 / 0.25: 44.7 / 0.375: 46.5 / \textbf{0.50: 48.2} / 0.625: 46.7 / 0.75: 45.1 / 0.875: 43.8 \\
Cold start         & conservative 46.4 / \textbf{calibration 48.2} \\
Prefix cut         & exact-token 46.8 / \textbf{sentence 48.2} \\
Control granularity& PrefixRL 35.9 / PrefixRL+anneal 38.7 / global decay 36.8 / base-only 45.3 / \textbf{base + offsets 48.2} \\
Prefix source      & off-policy (rejection-sampled) 46.2 / \textbf{reference 48.2} \\
Advantage norm.    & std-normalized 44.1 / \textbf{un-normalized 48.2} \\
Annealing envelope & none (eval @$\rho{=}0$) 44.8 / \textbf{delayed ($w{=}0.8$) 48.2} \\
Group size $G$     & 4: 43.6 / 8: 48.2 / \textbf{16: 48.7} \\
\bottomrule
\end{tabular}
\end{table}

\subsection{Analysis}
\label{sec:analysis}
\begin{figure}[t]
\centering
\begin{tikzpicture}
\begin{axis}[adaaxis, width=0.62\linewidth, height=4.3cm,
  ybar, bar width=4pt,
  xlabel={per-sample success rate $\kg{}$}, ylabel={\% of problems},
  xmin=-0.07, xmax=1.07, ymin=0, ymax=56,
  xtick={0,0.25,0.5,0.75,1},
  legend style={at={(0.14,0.98)}, anchor=north west, font=\scriptsize, fill=white, fill opacity=0.9, text opacity=1},
  legend cell align=left,
]
\fill[deadred!10] (axis cs:-0.07,0) rectangle (axis cs:0.06,56);
\fill[deadred!10] (axis cs:0.94,0) rectangle (axis cs:1.07,56);
\node[anchor=north, font=\scriptsize, deadred!85] at (axis cs:0.0,55) {dead};
\node[anchor=north, font=\scriptsize, deadred!85] at (axis cs:1.0,55) {dead};
\draw[adagold, dashed, line width=0.9pt] (axis cs:0.5,0) -- (axis cs:0.5,56);
\node[anchor=north, font=\scriptsize, adagold] at (axis cs:0.5,55) {$\tau$};
\addplot[fill=prefblue!45, draw=prefblue, area legend] coordinates {(0,8.0)(0.125,3.4)(0.25,2.6)(0.375,2.8)(0.5,3.6)(0.625,5.2)(0.75,8.4)(0.875,14.0)(1.0,52.0)};
\addlegendentry{PrefixRL}
\addplot[fill=adagold!75, draw=adagold, area legend] coordinates {(0,4.1)(0.125,7.2)(0.25,11.4)(0.375,18.9)(0.5,23.7)(0.625,19.3)(0.75,9.8)(0.875,4.2)(1.0,1.4)};
\addlegendentry{\method{} (ours)}
\end{axis}
\end{tikzpicture}
\caption{Distribution of per-sample \kg{} at mid-training (Qwen3-1.7B). Base control with per-sample offsets concentrates mass near the $\tau\!=\!0.5$ set-point, where signal is maximal; a fixed prefix has drifted toward saturation, with most problems piled at $\kg{}\!=\!1$ and a hard core stuck at $\kg{}\!=\!0$, since its lengths were tuned once at the start; most of its rollouts thus fall in the two zero-signal modes.}
\label{fig:dist}
\end{figure}
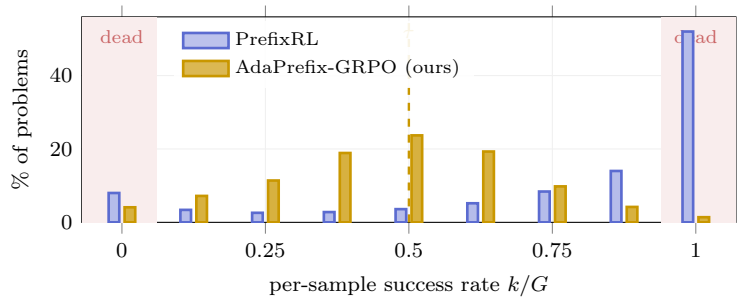

\paragraph{The controller tracks the set-point.} The closed loop holds the batch-mean \kg{} at $\tau\!=\!0.5$ within a few update intervals and keeps it there as the policy improves (\Cref{fig:mechanism}b) and keeps the \emph{whole distribution} of per-sample \kg{} concentrated near $\tau$, whereas fixed and global prefixes drift toward saturation, piling mass at $\kg{}\!=\!1$ while a hard core stays at $0$ (\Cref{fig:dist}). Tracking difficulty cohorts, their $\rho_x$ and realized \kg{} are driven toward the tracked mean by the global controller plus static offsets and anneal cleanly to $\rho=0$ (\Cref{fig:trajectories}). This is the mechanism that DAPO approximates with dynamic sampling, oversampling to discard all-correct/all-wrong groups~\citep{yu2025dapo}, except that prefix control creates signal on problems no amount of resampling can rescue.

\paragraph{Why prefixing transfers to no prefix.} Training on a partial solution still requires the model to generate the remaining reasoning and be graded on the full trace, so it practices the same sub-skills it must later chain unaided; as prefixes anneal away, those skills are exercised from progressively earlier states. This matches the \emph{back-generalization} PrefixRL reports~\citep{prefixrl2026}: held-out (no-prefix) pass@1 rises throughout training (\Cref{fig:teaser}), not only once prefixes vanish. Three mechanisms plausibly explain why the train/test distribution shift one might expect does not materialize. First, the loss is masked on prefix tokens: the model never imitates the (possibly off-distribution) reference text and is only reinforced on its own successful continuations, so every token that receives gradient is on-policy and no \emph{imitation} drift accumulates (conditioning still shifts the \emph{state} distribution; the final $\rho=0$ pass rates below are the direct check that this shift does not persist). Second, a prefix changes only the start state of the episode, not the policy being learned: as in reverse-curriculum RL~\citep{xi2024r3}, the sub-skills reinforced on the remainder (algebraic manipulation, case analysis, self-verification) are the same ones a from-scratch solution must chain, and they transfer across starting positions. Third, the controller itself performs a soft anneal: as the policy improves, every $\kappa$ curve shifts left and the loop lowers $\rho_x$ to keep \kg{} at $\tau$ (\Cref{fig:trajectories}b), so the distribution of training start states drifts toward unprefixed problems \emph{on its own}; the envelope merely forces the final step to exactly $\rho=0$. As the direct test of consolidation, the final $\rho=0$ policy solves $\sim$17\% of the full originally-dead set at pass@1 ($\sim$22\% at pass@16), evidence that the gains persist without prefixes.

\paragraph{Monotonicity in practice.} Across problems the response $\kappa(\rho)$ violates monotonicity on $\sim$6\% of measured $(\rho,\kappa)$ pairs (mostly long, off-distribution reference prefixes), and on those the controller falls back to bracketing without diverging; the mid-training snapshot in \Cref{fig:dist} shows the resulting concentration of per-sample \kg{} near $\tau$ on raw (un-smoothed) quantized rates.

\paragraph{Why traces shorten.} Because each sample is held near a moderate \kg{}, positive advantages reward \emph{efficient} correct completions rather than long, lucky explorations; combined with annealing, the policy converges to shorter unaided ($\rho=0$) solutions than vanilla GRPO, which only finds reward through length-heavy exploration. (We treat this as a hypothesis to verify with length-conditioned analyses at $\rho=0$.)

\section{Conclusion}
\label{sec:conclusion}
GRPO discards its hardest problems because zero within-group reward variance yields zero gradient. Prefix length is a continuous, per-sample dial that can place any problem at almost any reachable success rate. Conceptually this removes GRPO's \emph{sampling} barrier, the dead zone in which hard problems yield no signal under a finite rollout budget, and shifts the binding constraint to the model's own \emph{capacity}: with a controllable signal, whether a problem is learnable turns on whether the model can \emph{complete} it given enough scaffolding, not on whether it can already solve it unaided. Prefixing cannot exceed that capacity ceiling or rescue a reference path the model is unable to follow, but within it, difficulty is no longer a hard cutoff on trainability. The contribution of \method{} is to \emph{control} that dial in closed loop toward the intermediate, signal-maximizing target \kg{}~\citep{sweet1,sweet2} and anneal it away, rather than fixing it in advance. At matched training compute this extracts signal where GRPO got none, beats a fixed-prefix baseline, and yields shorter solutions, with the largest gains on the smallest models.

\paragraph{Limitations.} \method{} needs reference (or off-policy) solutions and a reliable, near-binary verifier; we evaluate only on math reasoning and do not test noisy or non-verifiable domains, where behavior may differ. We adopt the intermediate ($\approx\!0.5$) set-point from prior work~\citep{sweet1,sweet2} and locate it empirically; we do not derive the exact per-sample optimum under reward noise. A subtler caveat: a $50\%$ success rate on the short \emph{remainder} of a heavily prefixed problem exercises credit assignment only over that remainder, which may be an easier learning signal than $50\%$ on a from-scratch trajectory; annealing is what forces the harder, full-length signal back in, and disentangling the two is future work. The offset map $\delta$ is a hand-tuned static heuristic: a sample whose cold-start difficulty is mis-estimated keeps a biased prefix that the global loop cannot correct individually, and upgrading the offset to a genuine per-sample second loop would require denser per-sample \kg{} estimates than $G{=}8$ with sparse revisits provides. The cold-start calibration and difficulty probe, while one-time and amortized across runs, are not free; cheap or training-free per-sample difficulty estimation is an interesting direction for future work. The prefix source also matters (reference paths can be off-distribution), though our ablation suggests the controller is largely orthogonal to it. Small benchmarks (AIME, AMC) carry wide intervals, so we rely on several benchmarks and avg@8 sampling rather than any single number. Underlying per-step data and training code are not released.

\appendix
\section{Measurement Details}
\label{app:measurement}
Quantities quoted in the main text are measured as follows. \emph{Monotonicity violation rate} ($\sim$6\%, Section~\ref{sec:analysis}): the fraction of adjacent $(\rho,\kappa(\rho))$ pairs with $\Delta\kappa<0$, pooled over the calibration sweep (512 problems $\times$ 4 adjacent grid intervals $=$ 2{,}048 pairs) and all subsequent controller reads. \emph{Secant fallback rate} ($\sim$17\%, Section~\ref{sec:algorithm}): 4 of the 24 controller updates ($\sim$240 steps, $N{=}10$) triggered the bisection fallback, 3 of them in the first third of training. \emph{Prefill overhead} ($<$15\%, Section~\ref{sec:method}): prefix tokens account for 14.6\% of all training-time tokens (prefill $+$ generated), measured over the full \method{} run.

\bibliographystyle{plainnat}

\begin{thebibliography}{99}
\bibitem[Guo et al.(2025)]{guo2025deepseek} DeepSeek-AI. DeepSeek-R1: Incentivizing Reasoning Capability in LLMs via Reinforcement Learning. arXiv:2501.12948, 2025.
\bibitem[Shao et al.(2024)]{shao2024deepseekmath} Z.\ Shao et al. DeepSeekMath: Pushing the Limits of Mathematical Reasoning in Open Language Models (GRPO). 2024.
\bibitem[Setlur et al.(2026)]{prefixrl2026} A.\ Setlur, Z.\ Wang, A.\ Cohen, P.\ Rashidinejad, S.\ M.\ Xie. Reuse your FLOPs: Scaling RL on Hard Problems by Conditioning on Very Off-Policy Prefixes (PrefixRL). arXiv:2601.18795, 2026.
\bibitem[Qu et al.(2026)]{pope2026} Y.\ Qu, A.\ Setlur, V.\ Smith, R.\ Salakhutdinov, A.\ Kumar. POPE: Learning to Reason on Hard Problems via Privileged On-Policy Exploration. arXiv:2601.18779, 2026.
\bibitem[Huang et al.(2025)]{prefixrft2025} Z.\ Huang, T.\ Cheng, Z.\ Qiu, Z.\ Wang, Y.\ Xu, E.\ M.\ Ponti, I.\ Titov. Blending Supervised and Reinforcement Fine-Tuning with Prefix Sampling (Prefix-RFT). arXiv:2507.01679, 2025.
\bibitem[Zhang et al.(2025)]{scafgrpo2025} X.\ Zhang, S.\ Wu, Y.\ Zhu, H.\ Tan, S.\ Yu, Z.\ He, J.\ Jia. Scaf-GRPO: Scaffolded Group Relative Policy Optimization for Enhancing LLM Reasoning. arXiv:2510.19807, 2025.
\bibitem[Huang et al.(2025)]{hintgrpo2025} Q.\ Huang, W.\ Dai, J.\ Liu, W.\ He, H.\ Jiang, M.\ Song, J.\ Chen, C.\ Yao, J.\ Song. Boosting MLLM Reasoning with Text-Debiased Hint-GRPO. ICCV, 2025. arXiv:2503.23905.
\bibitem[Bae et al.(2025)]{sweet1} S.\ Bae, J.\ Hong, M.\ Y.\ Lee, H.\ Kim, J.\ Nam, D.\ Kwak. Online Difficulty Filtering for Reasoning Oriented Reinforcement Learning. arXiv:2504.03380, 2025.
\bibitem[Gao et al.(2025)]{sweet2} Z.\ Gao, J.\ Kim, W.\ Sun, T.\ Joachims, S.\ Wang, R.\ Y.\ Pang, L.\ Tan. Prompt Curriculum Learning for Efficient LLM Post-Training. arXiv:2510.01135, 2025.
\bibitem[Xi et al.(2024)]{xi2024r3} Z.\ Xi et al. Training Large Language Models for Reasoning through Reverse Curriculum Reinforcement Learning (R3). ICML, 2024. arXiv:2402.05808.
\bibitem[Florensa et al.(2018)]{florensa2018} C.\ Florensa, D.\ Held, X.\ Geng, P.\ Abbeel. Automatic Goal Generation for Reinforcement Learning Agents. ICML, 2018.
\bibitem[Schulman et al.(2017)]{schulman2017ppo} J.\ Schulman, F.\ Wolski, P.\ Dhariwal, A.\ Radford, O.\ Klimov. Proximal Policy Optimization Algorithms. arXiv:1707.06347, 2017.
\bibitem[Ahmadian et al.(2024)]{ahmadian2024rloo} A.\ Ahmadian et al. Back to Basics: Revisiting REINFORCE-Style Optimization for Learning from Human Feedback in LLMs (RLOO). ACL, 2024.
\bibitem[Ouyang et al.(2022)]{ouyang2022instructgpt} L.\ Ouyang et al. Training Language Models to Follow Instructions with Human Feedback (InstructGPT). NeurIPS, 2022.
\bibitem[Lambert et al.(2024)]{lambert2024tulu} N.\ Lambert et al. T\"ulu 3: Pushing Frontiers in Open Language Model Post-Training. arXiv:2411.15124, 2024.
\bibitem[Yu et al.(2025)]{yu2025dapo} Q.\ Yu et al. DAPO: An Open-Source LLM Reinforcement Learning System at Scale. arXiv:2503.14476, 2025.
\bibitem[Liu et al.(2025)]{liu2025drgrpo} Z.\ Liu, C.\ Chen, W.\ Xu, P.\ Qi, T.\ Pang, C.\ Du, W.\ S.\ Lee, M.\ Lin. Understanding R1-Zero-Like Training: A Critical Perspective (Dr.~GRPO). arXiv:2503.20783, 2025.
\bibitem[Zelikman et al.(2022)]{zelikman2022star} E.\ Zelikman, Y.\ Wu, J.\ Mu, N.\ Goodman. STaR: Bootstrapping Reasoning with Reasoning. NeurIPS, 2022.
\bibitem[Dong et al.(2023)]{dong2023raft} H.\ Dong et al. RAFT: Reward rAnked FineTuning for Generative Foundation Model Alignment. TMLR, 2023.
\bibitem[Chu et al.(2025)]{chu2025sft} T.\ Chu et al. SFT Memorizes, RL Generalizes: A Comparative Study of Foundation Model Post-Training. arXiv:2501.17161, 2025.
\bibitem[Cui et al.(2025)]{cui2025entropy} G.\ Cui et al. The Entropy Mechanism of Reinforcement Learning for Reasoning Language Models. arXiv:2505.22617, 2025.
\bibitem[Yan et al.(2025)]{luffy2025} J.\ Yan et al. Learning to Reason under Off-Policy Guidance (LUFFY). arXiv:2504.14945, 2025.
\bibitem[Ma et al.(2025)]{relift2025} L.\ Ma, H.\ Liang, M.\ Qiang, L.\ Tang, X.\ Ma, Z.\ H.\ Wong, et al. Learning What Reinforcement Learning Can't: Interleaved Online Fine-Tuning for Hardest Questions (ReLIFT). arXiv:2506.07527, 2025.
\bibitem[Bengio et al.(2009)]{bengio2009curriculum} Y.\ Bengio, J.\ Louradour, R.\ Collobert, J.\ Weston. Curriculum Learning. ICML, 2009.
\bibitem[Kumar et al.(2010)]{kumar2010selfpaced} M.\ P.\ Kumar, B.\ Packer, D.\ Koller. Self-Paced Learning for Latent Variable Models. NeurIPS, 2010.
\bibitem[Yang et al.(2025)]{yang2025qwen3} A.\ Yang et al. Qwen3 Technical Report. arXiv:2505.09388, 2025.
\bibitem[Gemma Team(2026)]{gemma2026} Gemma Team, Google DeepMind. Gemma 4 Technical Report. arXiv:2607.02770, 2026.
\bibitem[Qwen Team(2026)]{qwen35} Qwen Team. Qwen3.5: Towards Native Multimodal Agents. 2026.
\bibitem[He et al.(2025)]{deepmath2025} Z.\ He et al. DeepMath-103K: A Large-Scale, Challenging, Decontaminated, and Verifiable Mathematical Dataset. arXiv:2504.11456, 2025.
\bibitem[Hu et al.(2022)]{hu2022lora} E.\ Hu et al. LoRA: Low-Rank Adaptation of Large Language Models. ICLR, 2022.
\bibitem[von Werra et al.(2020)]{vonwerra2020trl} L.\ von Werra et al. TRL: Transformer Reinforcement Learning. \url{https://github.com/huggingface/trl}, 2020.
\bibitem[Kwon et al.(2023)]{kwon2023vllm} W.\ Kwon et al. Efficient Memory Management for Large Language Model Serving with PagedAttention (vLLM). SOSP, 2023.
\bibitem[Snell et al.(2024)]{snell2024} C.\ Snell, J.\ Lee, K.\ Xu, A.\ Kumar. Scaling LLM Test-Time Compute Optimally Can Be More Effective than Scaling Model Parameters. arXiv:2408.03314, 2024.
\bibitem[Hendrycks et al.(2021)]{hendrycks2021math} D.\ Hendrycks et al. Measuring Mathematical Problem Solving with the MATH Dataset. NeurIPS, 2021.
\bibitem[Cobbe et al.(2021)]{cobbe2021gsm8k} K.\ Cobbe et al. Training Verifiers to Solve Math Word Problems (GSM8K). arXiv:2110.14168, 2021.
\end{thebibliography}

\end{document}